\documentclass[sigconf]{acmart}

\usepackage{booktabs} % For formal tables

\usepackage[acronym]{glossaries}
\usepackage[automake]{glossaries-extra}
\usepackage[colorinlistoftodos]{todonotes}
\usepackage{subcaption}
\usepackage{booktabs}
\hypersetup{draft}

\newabbreviation{ei}{EI}{expected improvement}
\newabbreviation{pm}{PV}{predicted value}
\newabbreviation{smbo}{SBO}{surrogate-based optimization}

\setcopyright{none}
\acmDOI{10.1145/nnnnnnn.nnnnnnn} % To be updated after completing copyright process
\acmISBN{978-x-xxxx-xxxx-x/YY/MM} % To be updated after completing copyright process
\acmConference[GECCO '20]{The Genetic and Evolutionary Computation Conference 2020}{July 8--12, 2020}{Cancun, Mexico}
\acmYear{2020}
\copyrightyear{2020}

\acmPrice{15.00}

\makeglossaries

\begin{document}
\title{Expected Improvement versus Predicted Value in Surrogate-Based Optimization}
%\title{Should Expected Improvement be the Default in Surrogate Model-Based Optimization?}
%\title{Should Expected Improvement be the Default Infill Criterion in Surrogate Model-Based Optimization?}
%\title{Should Expected Improvement be the Default in Bayesian Optimization?}
%\title{Should Expected Improvement be the Default Acquisition Function in Bayesian Optimization?}
%\title{Testing/Evaluating Infill Criteria for Surrogate Model-Based Optimization}
\author{Frederik Rehbach, Martin Zaefferer, Boris Naujoks,  Thomas Bartz-Beielstein}
\affiliation{%
  \institution{Institute for Data Science, Engineering, and Analytics\\TH Köln, Gummersbach, Germany}
}
\email{firstname.surname@th-koeln.de}

\renewcommand{\shortauthors}{Rehbach et al.}

\begin{abstract}
Surrogate-based optimization relies on so-called infill criteria (acquisition functions) to decide which point to evaluate next.
When Kriging is used as the surrogate model of choice (also called Bayesian optimization), 
one of the most frequently chosen criteria is expected improvement.
We argue that the popularity of expected improvement largely relies on its theoretical properties rather than empirically validated performance.
Few results from the literature show evidence, 
that under certain conditions, expected improvement may perform worse than something as simple as the predicted value of the surrogate model.
We benchmark both infill criteria in an extensive empirical study on the `BBOB' function set. 
This investigation includes a detailed study of the impact of problem dimensionality on algorithm performance.
The results support the hypothesis that exploration loses importance with increasing problem dimensionality.
A statistical analysis reveals that the purely exploitative search with the predicted value criterion performs better on most problems of five or higher dimensions.
Possible reasons for these results are discussed.
In addition, we give an in-depth guide for choosing the infill criteria based on prior knowledge about the problem at hand, its dimensionality, and the available budget. 
\end{abstract}




\maketitle

% !TeX root = rehb20a.tex
\section{Introduction}
Many real-world optimization problems require significant resources for each evaluation of a given candidate solution.
For example, evaluations might require material costs for laboratory experiments or computation time for extensive simulations.
In such scenarios, the available budget of objective function evaluations is often severely limited to a few tens or hundreds of evaluations.

One standard method to efficiently cope with such limited evaluation budgets is \gls{smbo}.
Surrogate models of the objective function are based on data-driven models,
which are trained using only a relatively small set of observations.
Predictions from these models partially replace expensive objective function evaluations.
Since the surrogate model is much cheaper to evaluate than the original objective function, an extensive search becomes feasible.
In each \gls{smbo} iteration, one new candidate solution is proposed by the search on the surrogate.
The candidate is evaluated on the expensive function and the surrogate is updated with the new data.
This process is iterated until the budget of expensive evaluations is depleted.

A frequently used surrogate model is Kriging (also called Gaussian process regression).
In Kriging, a distance-based correlation structure is determined with the observed data \cite{Forr08a}. 
The search for the next candidate solution on the Kriging model is guided by a so-called infill criterion or acquisition function.

A straightforward and simple infill criterion is the \gls{pm}.
The \gls{pm} of a Kriging model is an estimation or approximation of the objective function at a given point in the search space \cite{Forr08a}.
If the surrogate exactly reproduces the expensive objective function, then
optimizing the \gls{pm} yields the global optimum of the expensive objective function. 
In practice, especially in the early iterations of the \gls{smbo} procedure, the prediction of the surrogate will be inaccurate.
Many infill criteria do not only consider the raw predicted function value but also try to improve the model quality with each new iteration.
This is achieved by suggesting new solutions in promising but unknown regions of the search space.
In essence, such criteria attempt to simultaneously improve the local approximation quality of the optimum as well as the
global prediction quality of the surrogate.

The \gls{ei} criterion is considered as a standard method for this purpose~\cite{Jone98a}.
\gls{ei} makes use of the internal uncertainty estimate provided by Kriging.
The \gls{ei} of a candidate solution increases if the predicted value or the estimated uncertainty of the model rises.

The optimization algorithm might converge to local optima, if solely the \gls{pm} is used as an infill criterion.
In the most extreme case, if the \gls{pm} suggests a solution that is equal to an already known solution, the
algorithm might not make any progress at all, instead repeatedly suggesting the exact same solution~\cite{Forr08a}.
Conversely, \gls{ei} can yield a guarantee for global convergence, and convergence rates can be analyzed analytically~\cite{Bull11a,Wess17a}.
As a result, \gls{ei} is frequently used as an infill criterion. 

We briefly examined 24 software frameworks for \gls{smbo}. 
We found 15 that use \gls{ei} as a default infill criterion, three that use \gls{pm}, five that use criteria based on 
lower or upper confidence bounds \cite{Auer02a}, and a single one that uses a portfolio strategy (which includes confidence bounds and \gls{ei}, but not \gls{pm}).
An overview of the surveyed frameworks is included in the supplementary material. 
Seemingly, \gls{ei} is firmly established. 

Despite this, some criticism can be found in the literature.
Wessing and Preuss point out that the convergence proofs are of theoretical interest but have no relevance in practice~\cite{Wess17a}.
The most notable issue in this context is the limited evaluation budget that is often imposed on \gls{smbo}  algorithms.
Under these limited budgets, the asymptotic behavior of the algorithm can become meaningless.
Wessing and Preuss show results where a model-free CMA-ES often outperforms \gls{ei}-based EGO within a few hundred function evaluations~\cite{Wess17a}.

A closely related issue is that many infill criteria, including \gls{ei}, define a static trade-off between
exploration and exploitation~\cite{Gins10a,Lam16a,Wang18a,Wang19a}. 
This may not be optimal. Intuitively, explorative behavior may be of more interest in the beginning, rather than close to the end of an optimization run.

At the same time, benchmarks that compare \gls{ei} and \gls{pm} are far and few between.
While some benchmarks investigate the performance of \gls{ei}-based algorithms, they rarely
compare to a variant based on \gls{pm}  (e.g., \cite{Osbo10a,Snoe12a,Hern14a,Nguy16b}).
Other authors suggest portfolio strategies that combine multiple criteria.
Often, tests of these strategies do not include the \gls{pm} in the portfolio (e.g., \cite{Hoff11a,Vasc19a}). 

We would like to discuss two exceptions. Firstly, Noe and Husmeier~\cite{Noe18a}
do not only suggest a new infill criterion and compare it to \gls{ei}, but they also
compare it to a broad set of other criteria, importantly including \gls{pm}.
In several of their experiments, \gls{pm} seems to perform better than \gls{ei}, depending
on the test function and the number of evaluations spent.
Secondly, a more recent work by De Ath et al.~\cite{DeAt19a} reports similar observations.
They observe that``\textit{Interestingly, Exploit, which always samples from the best mean surrogate prediction is competitive for most of the high dimensional problems}''~\cite{DeAt19a}. 
Conversely, they observe better performances of explorative approaches on a few, low-dimensional problems.

In both studies, the experiments cover a restricted set of problems where the dimensionality
of each specific test function is usually fixed. 
The influence of test scenarios, especially in terms of the search space dimension, remains to be clarified.
It would be of interest, to see how increasing or decreasing the dimension affects results
for each problem.

Hence, we raise two tightly connected research questions:
\begin{itemize}
\item[\textbf{RQ-1}] Can distinct scenarios be identified where \gls{pm} outperforms \gls{ei} or vice versa? 
\item[\textbf{RQ-2}] Is \gls{ei} a reasonable choice as a default infill criterion? 
\end{itemize}

Here, scenarios comprehend the whole search framework, such as features of the problem landscape,
dimension of the optimization problem, or the allowed evaluation budget. 
In the following, we describe experiments that investigate these research questions. 
We discuss the results based on our observations and a statistical analysis. 

\section{Experiments}
\subsection{Performance Measurement}
Judging the behavior of an algorithm requires an appropriate performance measure.
The choice of this measure depends on the context of our investigation.
A necessary pretext for our experiments is the limited budget of objective evaluations we allow for
focusing on algorithms for expensive optimization problems.
Since the objective function causes the majority of the optimization cost in such a scenario, a typical goal is to find the best possible candidate solution within a given amount of evaluations. 

For practical reasons, we have to specify an upper limit of function evaluations for our experiments. 
However, for our analysis, we evaluate the performance measurements for each iteration. 
Readers who are only interested in a scenario with a specific fixed budget can, therefore, just consider the results of the respective iteration.

\subsection{Algorithm Setup}
\begin{table*}[!h]
	\centering
	\caption{Overview of the 24 BBOB test functions including some of their landscape features.}
	\label{tab:BBOB}
	\begin{tabular}{l|l|l}
		ID & Name                & Specific Features          		 	 \\ \hline
		1  & Sphere              & unimodal, separable, symmetric  	  \\
		2  & Ellipsoidal         & unimodal, separable, high conditioning  \\
		3  & Rastrigin           & multimodal, separable, regular/symmetric structure \\
		4  & Büche-Rastrigin     & multimodal, separable, asymmetric structure          	\\
		5  & Linear Slope        & unimodal, separable             	\\
		6  & Attractive Sector   & unimodal, low/moderate conditioning , asymmetric structure    	\\
		7  & Step Ellipsoidal    & unimodal, low/moderate conditioning, many plateaus     	\\
		8  & Rosenbrock          & unimodal/bimodal depending on dimension, low/moderate conditioning    	\\
		9  & Rosenbrock, rotated & unimodal/bimodal depending on dimension, low/moderate conditioning    \\
		10 & Ellipsoidal         & unimodal, high conditioning \\
		11 & Discus              & unimodal, high conditioning  \\
		12 & Bent Cigar          & unimodal, high conditioning \\ 
		13 & Sharp Ridge              &unimodal, high conditioning   \\
		14 & Different Powers         &unimodal, high conditioning           \\
		15 & Rastrigin                & multimodal, non-separable, low conditioning, adequate global structure \\
		16 & Weierstrass              & multimodal (with several global otpima), repetitive, adequate global structure \\
		17 & Schaffers F7             & multimodal, low conditioning, adequate global structure \\
		18 & Schaffers F7             & multimodal, moderate conditioning, adequate global structure \\
		19 & Griwank-Rosenbrock       & multimodal, adequate global structure \\
		20 & Schwefel                 & multimodal, weak global structure (in the optimal, unpenalized region)     \\
		21 & Gallagher G.101-me Peaks & multimodal, low conditioning, weak global structure     \\
		22 & Gallagher G. 21-hi Peaks & multimodal, moderate conditioning, weak global structure     \\
		23 & Katsuura                 & multimodal, weak global structure     \\
		24 & Lunacek bi-Rastrigin     & multimodal, weak global structure    
	\end{tabular}
\end{table*}

One critical contribution to an algorithm's performance is its configuration and setup.
This includes parameters as well as the chosen implementation.

We chose the R-package `SPOT' \cite{Bart05a,SPOTv2.0.4} as an implementation for \gls{smbo}.
It provides various types of surrogate models, infill criteria, and optimizers.
To keep the comparison between the \gls{ei} and \gls{pm} infill criteria as fair as possible, they will be used with the same configuration in `SPOT'.
This mentioned configuration regards the optimizer for the model, the corresponding budgets, and lastly the kernel parameters for the Kriging surrogate. 
The configuration is explained in more detail in the following.

We allow each `SPOT' run to expend 300 evaluations of the objective function. 
This accounts for the assumption that \gls{smbo} is usually applied in scenarios with severely limited evaluation budgets.
This specification is also made due to practical reasons:
the runtime of training Kriging models increases exponentially with the number of data points. 
Hence, a significantly larger budget would limit the number of tests we could perform due to excessive algorithm runtimes. 
Larger budgets, for instance, 500 or 1\,000 evaluations, may easily lead to infeasible high runtimes
(depending on software implementation, hardware, and the number of variables). 
If necessary, this can be partially alleviated with strategies that reduce runtime complexity, such as clustered Kriging~\cite{VanS15a,Wang2017}. 
However, these strategies have their own specific impact on algorithm performance, which we do not want to cover in this work.

The first ten of the 300 function evaluations are spent on an initial set of candidate solutions. 
This set is created with Latin Hypercube sampling \cite{McKa79a}.
In the next 290 evaluations, a Kriging surrogate model is trained at each iteration. 
The model is trained with the `buildKriging' method from the `SPOT' package.
This Kriging implementation is based on work by Forrester et. al. \cite{Forr08a}, and uses an anisotropic kernel:
$$k(\mathbf{x},\mathbf{x}')=\exp\left(\sum_{i=1}^{d} -\theta_i |x_i-x'_i|^{p_i}\right).$$  
Here, $\mathbf{x}, \mathbf{x}' \in \mathbb{R}^d$  are candidate solution vectors, with dimension $d$ and $i=1,\ldots,d$. Furthermore, the kernel has parameters $\theta_i \in \mathbb{R}_+$ and $p_i \in [0.01,2]$.
During the model building, the nugget effect \cite{Forr08a}
is activated for numerical stability, and the respective parameter $\lambda$ is set to a maximum of $10^{-4}$.
Parameters $\theta_i$, $p_i$ and $\lambda$ are determined by Maximum Likelihood Estimation (MLE) via differential evolution \cite{Stor97a} (`optimDE').
The budget for the MLE is set to $500 \times t$ 
likelihood evaluations, where $t = 2d + 1$ is the number of model parameters optimized by MLE ($\theta$,$p$, $\lambda$).

The respective infill criterion is optimized, after the model training.
Again, differential evolution is used for this purpose. 
In this case, we use a budget of $1\,000 \times d$ evaluations of the model (in each iteration of the \gls{smbo} algorithm).
Generally, these settings were chosen rather generously to ensure high model quality and a well-optimized infill criterion.

To rule out side effects caused by potential peculiarities of the implementation, independent tests were performed with a different implementation based on the R-package `mlrmbo' \cite{Bisc17a,mlrMBOv1.1.2}.
Notably, the tests with `mlrmbo' employed a different optimizer (focus search instead of differential evolution) and a different Kriging implementation, based on  DiceKriging \cite{Rous12a}.

\subsection{Test Functions}
To answer the presented research questions, we want to investigate how different infill criteria for \gls{smbo} behave on a broad set of test functions.
We chose one of the most well-known benchmark suites in the evolutionary computation community: the Black-Box Optimization Benchmarking (BBOB)\footnote{https://coco.gforge.inria.fr/} suite \cite{Hans16a}. 
While the test functions in the BBOB suite themselves are not expensive to evaluate, 
we emulate expensive optimization by limiting the algorithms to only a few hundred evaluations.
The standard BBOB suite contains 24 noiseless single-objective test functions, divided into five groups by common features and landscape properties.
The global optima and landscape features of all functions are known. 
An overview of the 24 functions and their most important properties is given in Table \ref{tab:BBOB}.
Hansen et al.\/ present a detailed description ~\cite{Hans09a}.

Each of the BBOB test functions is scalable in dimensionality so that algorithms can be compared with respect to performance per dimension. 
Furthermore, each function provides multiple instances.
An instance is a rotated or shifted version of the original objective function.
All described experiments were run with a recent GitHub version\footnote{https://github.com/numbbo/coco}, v2.3.1 \cite{Hans19a}.
Each algorithm is tested on the 15 available standard instances of the BBOB function set. 

Preliminary experiments were run using the smoof test function set implemented in the `smoof' R-package~\cite{Boss17a,smoofv1.5.1}.\label{sec:smoofExplanation}
The smoof package consists of a total of more than 70 single-objective functions as well as 24 multi-objective functions.
Since our analysis will consider observations in relation to the scalable test function dimension, 
we do not further discuss the complete results of the (often non-scalable) smoof test function set.
Yet, considering the fixed dimensions in smoof, similar observations were made for both smoof and BBOB.

\section{Results}
\subsection{Convergence Plots} \label{sec:Converg}
\begin{figure*}[!h]
	\begin{subfigure}[c]{\textwidth}
    \includegraphics[width=\textwidth, page=3]{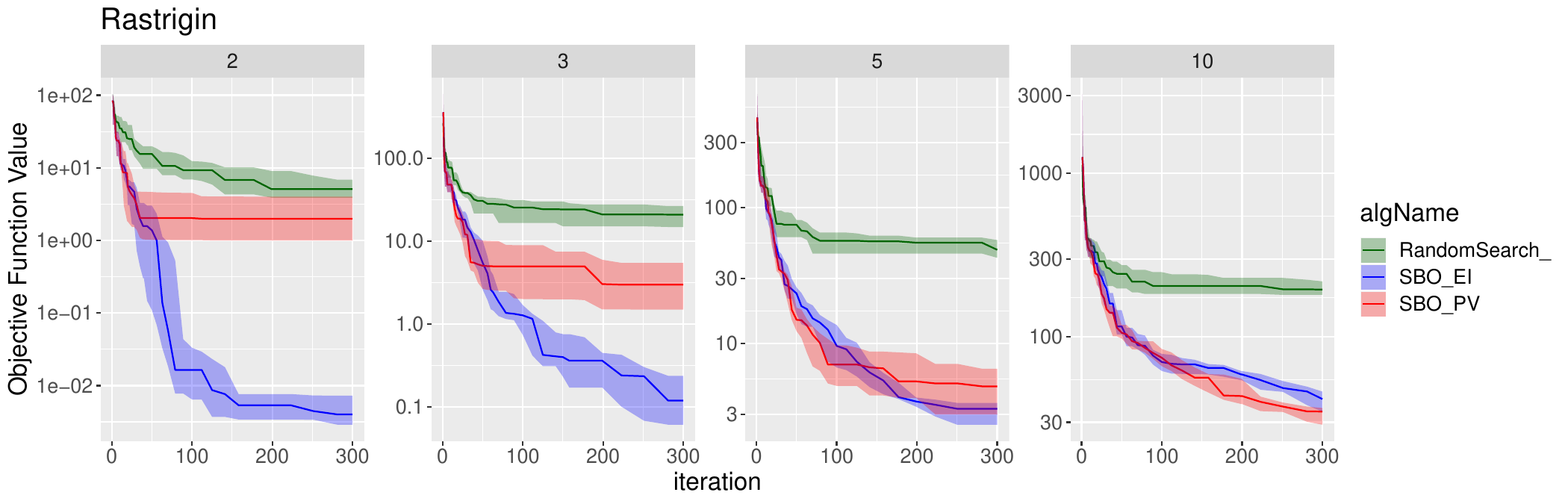}
    \subcaption{BBOB Function 3 - Separable Rastrigin}
  \end{subfigure}
  \begin{subfigure}[c]{\textwidth}
    \includegraphics[width=\textwidth, page=13]{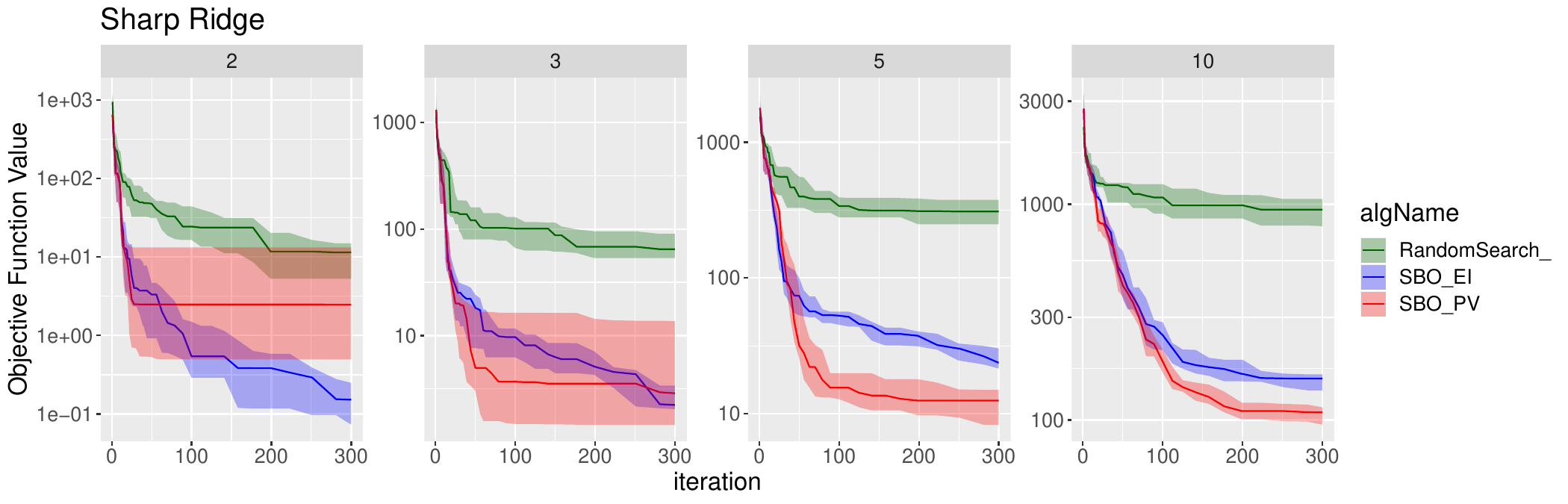}
    \subcaption{BBOB Function 13 - Sharp Ridge}
 	\end{subfigure}
 	\caption{Convergence plots of \gls{smbo}-\gls{ei} and \gls{smbo}-\gls{pm} on two of the 24 BBOB functions. Random search added as baseline. The experiments are run (from left to right) on 2,3,5, and 10 input-dimensions. The $y$-axis shows the best so far achieved objective function value on a logarithmic scale. The center line indicates the median of the algorithm repeats. The surrounding ribbon marks the lower and upper quartiles. $Y$-values measure differences from the global function optimum.}
 	\label{fig:convergence}
\end{figure*}
Our first set of results is presented in the form of convergence plots shown in Figure \ref{fig:convergence}. 
The figure shows the convergence of the \gls{smbo}-\gls{pm} and \gls{smbo}-\gls{ei} algorithm on two of the 24 BBOB functions.
These two functions (Rastrigin, Sharp Ridge) were chosen because they nicely represent the main characteristics and problems that can be observed for both infill criteria.
These characteristics will be explained in detail in the following paragraphs. 
A simple random search algorithm was included as a baseline comparison. 
The convergence plots of all remaining functions are available as supplementary material, also included are the results with the R-package `mlrmbo'.

As expected, the random search is largely outperformed on three to ten-dimensional function instances.
In the two-dimensional instances, the random search approaches the performance of \gls{smbo}-\gls{pm}, at least for
longer algorithm runtimes.
For both functions shown in Figure \ref{fig:convergence}, it can be observed that \gls{smbo}-\gls{ei} works as good or even better than \gls{smbo}-\gls{pm} on the two-, and three-dimensional function instances. 
As dimensionality increases, \gls{smbo}-\gls{pm} gradually starts to overtake and then outperform \gls{smbo}-\gls{ei}.

Initially, \gls{smbo}-\gls{pm} seems to have a faster convergence speed on both functions. 
Yet, this speedup comes at the cost of being more prone to getting stuck in local optima or sub-optimal regions. 
This problem is most obvious in the two- and three-dimensional instances, especially on the two-dimensional instance of BBOB function 3 (Separable Rastrigin).
Here, \gls{smbo}-\gls{ei} shows a similar convergence to \gls{smbo}-\gls{pm} on roughly the first 40 iterations.
However, after just 100 iterations, the algorithm appears to get stuck in local optima, reporting only minimal or no progress at all.
As Rastrigin is a highly multi-modal function, the results for \gls{smbo}-\gls{pm} are not too surprising.
At the same time, \gls{smbo}-\gls{ei} yields steady progress over all 300 iterations, exceeding the performance of \gls{smbo}-\gls{pm}.

Yet, this promising performance seems to change on higher dimensional functions.
On the five- and ten-dimensional function sets, the performance of \gls{smbo}-\gls{pm} is close to the one of \gls{smbo}-\gls{ei} early on.
Later in the run, \gls{smbo}-\gls{pm} outperforms \gls{smbo}-\gls{ei} noticeably. 
Neither algorithm shows a similar form of stagnation as was visible for \gls{smbo}-\gls{pm} in the lower dimensional test scenarios.

This behavior indicates that with increasing dimensionality, it is less likely for \gls{smbo} to get stuck in local optima. 
Therefore, the importance of exploration diminishes with increasing problem dimension. 
De Ath et al. reach a similar conclusion for the higher dimensional functions they considered ~\cite{DeAt19a}.
They argue that the comparatively small accuracy that the surrogates can achieve on high dimensional functions results in some internal exploration even for the strictly exploitative \gls{smbo}-\gls{pm}.
This is due to the fact that the estimated location for the function optimum might be far away from the true optimum.
In Section \ref{sec:caseStudy} this is covered in more detail. 
There, we investigate exactly how much exploration is done by each infill criterion.

\subsection{Statistical Analysis}
To provide the reader with as much information as possible in a brief format, the rest of this section will present data that was aggregated via a statistical analysis. 
We do not expect that our data is normal distributed.
For example, we know that we have a fixed lower bound on our performance values.
Also, our data is likely heteroscedastic (i.e., group variances are not equal).
Hence, common parametric test procedures that assume homoscedastic (equal variance), normal distributed data may be unsuited.

\begin{figure}[]
	\includegraphics[width=0.49\textwidth]{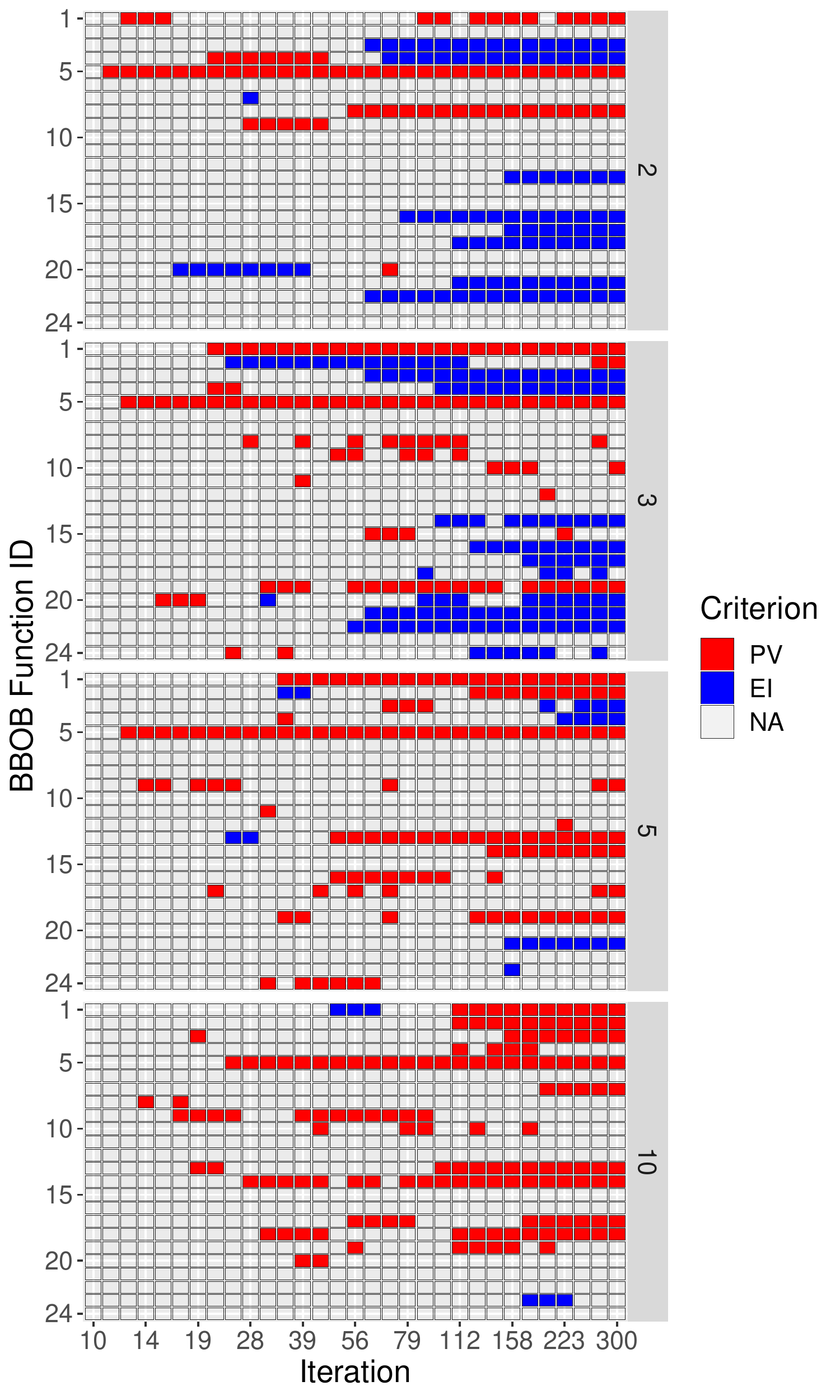}
	\caption{Statistical analysis of difference in performance between \gls{smbo}-\gls{ei} vs. \gls{smbo}-\gls{pm}. Y-Axis shows the BBOB-function index. 
		Colors indicate which infill criterion performed better in each case, based on the statistical test procedure. 
		Blank boxes indicates that there was no significant difference found.
	The results are presented for the 2,3,5 and 10 dimensional BBOB-function set as indicated in the gray bar to the right of each plot segment.
	The BBOB suite reports results in exponentially increasing intervals, thus the figure follows the same exponential scale on the x-axis. }
	\label{fig:dominations}
\end{figure}

Therefore, we apply non-parametric tests that make less assumptions about the underlying data, as suggested by Derrac et al.~\cite{Derr11a}.
We chose the Wilcoxon test, also known as Mann-Whitney test~\cite{Holl14a}, and use the test implementation from the base-R package `stats'.
Statistical significance is accepted if the corresponding p-values are smaller than $\alpha= 0.05$.
The statistical test is applied to the results of each iteration of the given algorithms. 
As the BBOB suite reports results in exponentially increasing intervals, the plot in Figure \ref{fig:dominations} follows the same exponential scale on the x-axis. 
The figure shows the aggregated results of all tests on all functions and iterations. 
Blue cells indicate that \gls{smbo}-\gls{ei} significantly outperformed \gls{smbo}-\gls{pm}, while red indicates the opposite result.
Uncolored cells indicate that there was no evidence for a statistically significant difference between the two competing algorithms.
The figure is further split by the input dimensionality (2,3,5,10) of the respective function, which is indicated on the right-hand side of each subplot. 

%In addition to discussing the results regarding specific functions and dimensionality, we will also consider the five function classes which are given in the BBOB suite: i) separable functions, ii) functions with low or moderate conditioning, iii) functions with high conditioning and unimodal, iv) multimodal functions with adequate global structure, and v) multimodal functions with weak global structure.
%On overview of which function belongs to which landscape group is given in Table \ref{tab:BBOB}.

We start with an overview of the two-dimensional results.
Initially, \gls{smbo}-\gls{pm} seems to perform slightly better than its competitor. 
Until roughly iteration 60, it shows statistical dominance on more functions than \gls{ei}.
Yet, given more iterations, EI performs well on more functions than \gls{pm}.
The performance of \gls{smbo}-\gls{pm} with less than 60 iterations, together with the convergence plots, indicates that \gls{smbo}-\gls{pm} achieves faster convergence rates.
Thus, \gls{smbo}-\gls{pm} is considered the more greedy algorithm.

This same greediness of \gls{smbo}-\gls{pm} may increasingly lead to stagnation when more than 60 iterations are performed.
Here, the \gls{smbo}-\gls{pm} algorithm is getting stuck at solutions that are sub-optimal or only locally optimal.
\gls{smbo}-\gls{ei} starts to outperform \gls{smbo}-\gls{pm}, overtaking it at roughly 70 iterations.
At the maximum budget of 300 function evaluations, \gls{smbo}-\gls{ei} outperforms \gls{smbo}-\gls{pm} on 8 of the 24 test functions.
\gls{smbo}-\gls{pm} only works best on 3 out of the 24 functions for the two-dimensional instances (Sphere, Linear Slope, Rosenbrock). 
Notably, those three functions are unimodal (at least for the two-dimensional case discussed so far).
It is not surprising to see the potentially more greedy \gls{smbo}-\gls{pm} perform well on unimodal functions.

A similar behavior is observed on the three-dimensional function set.
Initially, \gls{smbo}-\gls{pm} performs well on up to five functions, while \gls{smbo}-\gls{ei} only performs better on up to two functions. 
At around 85 iterations, \gls{smbo}-\gls{ei} again overtakes \gls{smbo}-\gls{pm} and then continues to perform well on more than eight functions up to the maximum budget of 300 iterations. 

Based on the observations for the two- and three-dimensional scenarios, 
one could assume that a similar pattern would be observed for higher-dimensional functions.
However, as previously discussed, it seems that \gls{smbo}-\gls{pm}'s convergence rate is less likely to stagnate with increasing problem dimensionality. 
On the five-dimensional functions, only three functions remain on which \gls{smbo}-\gls{ei} outperforms at the maximum given budget.
Namely: Rastrigin, Büche-Rastrigin, and Gallagher's Gaussian 101-me peaks function. 
All of which have a large number of local optima. 
Furthermore, Gallagher's function has little to no global structure, thus requiring a more explorative search.
On the other hand, \gls{smbo}-\gls{pm} performs better on up to 7 functions. 
This now also includes multimodal functions, hence functions that are usually not considered promising candidates for \gls{smbo}-\gls{pm}.

On the ten-dimensional function set, \gls{smbo}-\gls{ei} is outperformed on nearly all functions, with only two temporal exceptions. 
Only on the Sphere function and the Katsuura function, a statistically significant difference can be measured for a few iterations in favour of \gls{smbo}-\gls{ei}. 
\gls{smbo}-\gls{pm} performs significantly better on 9 out of the 24 functions.
Only on the function group with weak global structure, \gls{smbo}-\gls{pm} fails to produce significantly better performance.

Summarizing these results for separate function groups, it is noticeable that \gls{smbo}-\gls{ei} tends to work better on the multimodal functions.
Here, \gls{smbo}-\gls{ei} clearly excels on the two-, and three-dimensional instances.
On the functions with weak global structure, it continues to excel on the five-dimensional functions and is at least not outperformed on the ten-dimensional ones.

\gls{smbo}-\gls{ei} performs especially poor on `functions with low or moderate conditioning'.
Here, \gls{smbo}-\gls{pm} outperforms \gls{smbo}-\gls{ei} on at least as many functions as vice versa, independent of the budget and the problem dimensionality.
Generally, multimodality does not seem to require an explorative search methodology as long as the input dimensionality is high. 

\subsection{Case Study: Measuring Exploration}\label{sec:caseStudy}
\begin{figure*}[!h]
		\begin{subfigure}[c]{\textwidth}
		\includegraphics[width=\textwidth,page=3]{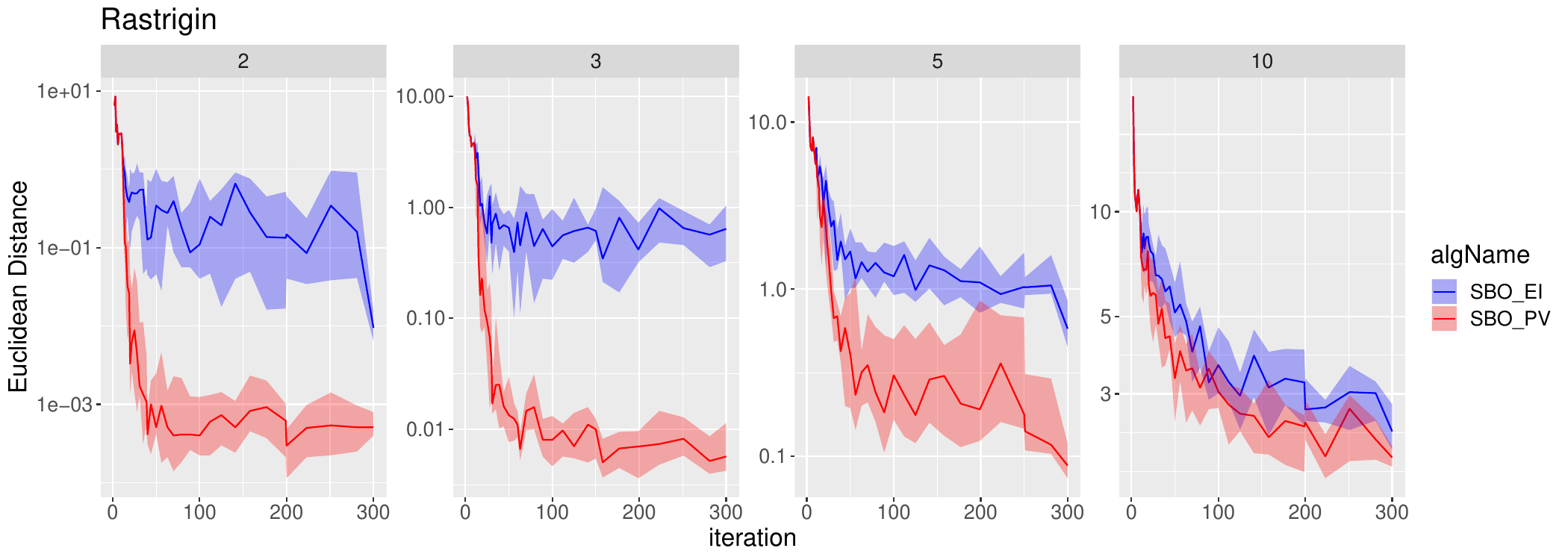}
		\subcaption{BBOB Function 3 - Separable Rastrigin}
	\end{subfigure}
	\begin{subfigure}[c]{\textwidth}
		\includegraphics[width=\textwidth,page=13]{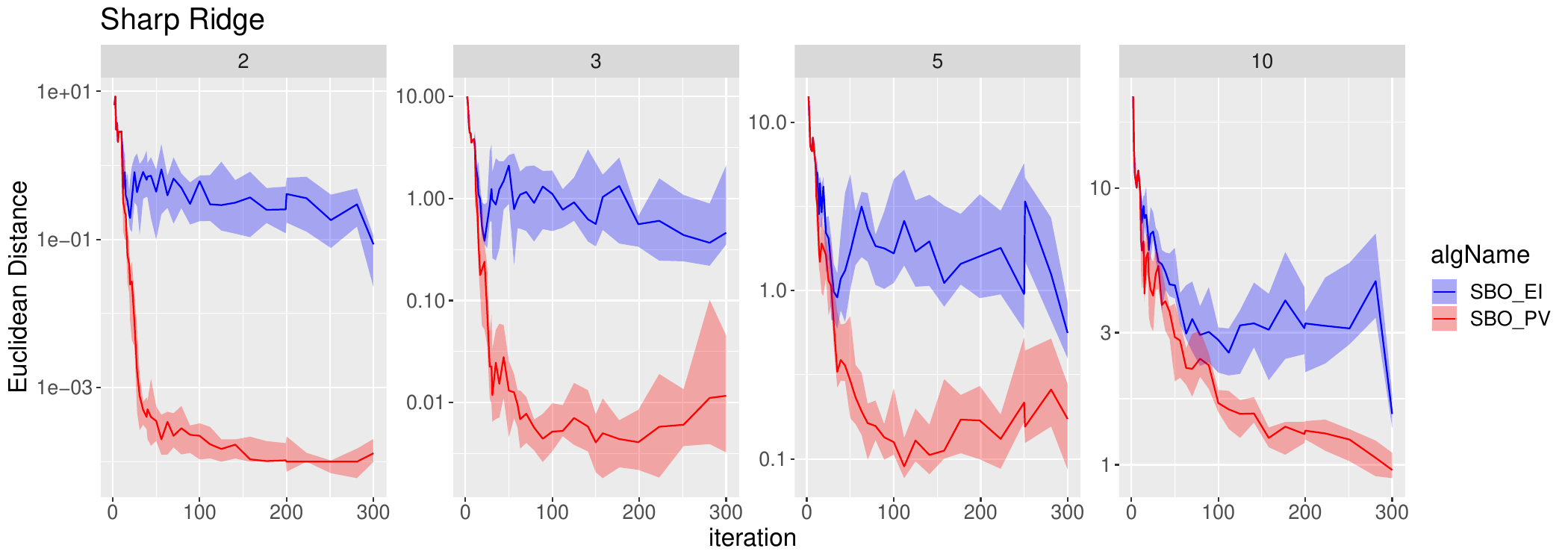}
		\subcaption{BBOB Function 13 - Sharp Ridge}
	\end{subfigure}
	\caption{The proposed candidate solutions' distance to its nearest neighbor in the set of known solutions. The $y$-axis shows the measured Euclidean distance. The experiments are run (from left to right) on 2,3,5, and 10 input-dimensions. The colored lines indicates the median of the distance, computed over the repeated runs of each algorithm. The surrounding ribbon marks the lower and upper quartiles.}
	\label{fig:distance}
\end{figure*}

The performance discussion in Section \ref{sec:Converg} largely relies on the idea that \gls{smbo}-\gls{pm} generally does less exploration than \gls{smbo}-\gls{ei}.
Multiple functions were observed on which the search with \gls{smbo}-\gls{pm} stagnates too early (cf. Section \ref{sec:Converg}). 
On the other hand, we argue that as long as \gls{smbo}-\gls{pm} does not get stuck in sub-optimal regions,
its greedy behavior leads to faster convergence and, thus, a more efficient search process.
This may especially be true if the budget is small in relation to the problem dimension.
In other words, if time is short, greedy behavior may be preferable.

To support this hypothesis, additional experiments were carried out to determine the amount of "exploration" each criterion does in different stages of the optimization runs.
For this purpose, we assume that exploration can be estimated as the Euclidean distance of a proposed candidate to its closest neighbor in the already evaluated set of candidate solutions.
Therefore, placing a new point close to a known point is regarded as an exploitative step.
Else, placing a new point far away from known points is considered as an explorative step.

This measure was applied to the previously discussed optimization runs on the BBOB function set.
The results for the two functions are shown in Figure \ref{fig:distance}.
Represented are the same functions as in the previous section for comparability. 
Results of the case study on all 24 BBOB functions are included in the supplementary material.

As discussed earlier, the performance of \gls{smbo}-\gls{pm} stagnated on the two- and three-dimensional instances of this function.
Hence, it is interesting to see that the measured distance becomes fairly small in those same cases.
While \gls{smbo}-\gls{ei} seems to make comparatively large search steps, \gls{smbo}-\gls{pm} seems to propose candidate solutions in the close vicinity of already known solutions.

On the higher-dimensional functions, the difference between \gls{smbo}-\gls{ei} and \gls{smbo}-\gls{pm} decreases.
On the Rastrigin function, both infill criteria reach similar levels on the ten-dimensional function. 
On the Sharp Ridge problem, a significant difference between the criteria remains. But the difference decreases with increasing problem dimension.
This supports the idea that the convergence speed of \gls{smbo}-\gls{pm} is generally higher due to less time being spent on exploring the search space. 

%\section{Conclusion}
\section{Discussion}
Before giving a final conclusion and an outlook on possible future work, we will reconsider the underlying research questions
of this investigation.

\textbf{RQ-1} Can distinct scenarios be identified where \gls{pm} outperforms \gls{ei} or vice versa? 

The large quantity of results measured on the BBOB and the smoof function sets (cf. Section \ref{sec:smoofExplanation}) allows identifying scenarios in
which either \gls{pm} or \gls{ei} perform best.

\begin{itemize}
	\item \textbf{Problem dimension:} 
	Considering the previously presented results, \gls{smbo}-\gls{ei} performs better on lower-dimensional functions, whereas \gls{smbo}-\gls{pm} excels on higher dimensional functions.
	If no further knowledge about a given optimization problem is available, then the benchmark results indicate that it is best to apply \gls{smbo}-\gls{ei} to functions with up to three dimensions.
	Problems with five or more input dimensions are likely best solved by \gls{smbo}-\gls{pm}.
	\item \textbf{Budget:} 
	The main drawback of an exploitative search (\gls{smbo}-\gls{pm}) is the likelihood of prematurely converging into a local optimum. 
	The larger the available budget of function evaluations, the more likely it is that \gls{smbo}-\gls{pm} will converge to a local optimum. 
	Yet, as long as \gls{smbo}-\gls{pm} is not stuck, it is more efficient in finding the overall best objective function value.
	Therefore, for many functions, there should be a certain \textit{critical budget} until which \gls{smbo}-\gls{pm} outperforms \gls{smbo}-\gls{ei}. 
	If the budget is larger, \gls{smbo}-\gls{ei} performs better.
	The results indicate that this \textit{critical budget} is increasing with input dimensionality.
	In the discussed benchmarks, \gls{smbo}-\gls{ei} started to excel after roughly 70 iterations on the two- and three-dimensional functions. 
	On the five-dimensional set, \gls{smbo}-\gls{ei} started excelling on some functions at around 200 iterations. 
	On the ten-dimensional, no signs of approaching such a \textit{critical budget} could be observed, indicating that it might lie far beyond (i.e. \(>\!\!>\) 300 evaluations) reasonable budgets for \gls{smbo}.
	\item \textbf{Modality:} 
	Lastly, any a priori knowledge regarding the landscape of the given function can be used to make a more informed decision.
	For this, the function classes and landscape features of the BBOB function should be considered.
	The greatest weakness of \gls{smbo}-\gls{pm} is to get stuck in local optima (or flat, sub-optimal regions).
	Hence, if it is known that a given function is fairly multimodal, then \gls{ei} may be a good choice.
	For simpler, potentially unimodal functions, we recommend using \gls{pm}.
	%Any more landscape features?
\end{itemize}

\textbf{RQ-2} Is \gls{ei} a reasonable choice as a default infill criterion? 

Since the best infill criterion changes depending on the optimization problem,
a simple answer to this question can not be given. 
We suggest that the default choice should be determined depending on
what is known about the use case. 

\begin{itemize}
	\item \textbf{Problem dimension:} 
	As discussed above, problem dimension is a crucial factor.
	Even when optimizing a black-box function, the input dimensionality of the function should be known a priori. 
	Therefore, being able to select a proper infill criterion based on input dimensionality should be applicable to most optimization tasks.
	The problem dimension is usually provided to the algorithm when the user specifies the search bounds.
	
	\item \textbf{Budget:}
	The available budget of function evaluations can be considered for the selection of a good infill criterion.
	As discussed above, the number of evaluations has a strong impact on whether the \gls{ei} or the \gls{pm} criterion
	performs better.
	However, the budget can not be determined automatically and will depend on fairly problem-specific
	knowledge. 
	In practice, the importance of this has to be communicated to practitioners.
	
	\item \textbf{Modality:} 
	Another impact factor is the modality of the search landscape. 
	For black-box problems, this is usually not known.
	To avoid premature convergence, a conservative choice would be to select
	\gls{ei} (as long as the problems are low-dimensional and the budget is relatively high).
	
	\item \textbf{Criterion Complexity:}
	Regardless of performance, there may be an additional argument for selecting 
	the \gls{pm} criterion: its simplicity.
	This has two consequences. 
	Firstly, it is easier to implement and compute. 
	Secondly, it is easier to explain, which may help to bolster the acceptance of
	\gls{smbo} algorithms in practice.
\end{itemize}

\section{Conclusion}
In conclusion, we observe that \gls{smbo}-\gls{ei} performs differently than often assumed.
Especially on higher-dimensional functions, \gls{smbo}-\gls{pm} seems to be the better choice,
despite (or because) of its fairly greedy search strategy.
Only on lower-dimensional, multimodal functions, \gls{smbo}-\gls{pm} is more likely to get stuck in local optima, at least
given the limited budgets under which our benchmark study was performed. 
In these cases, the largely explorative approach of \gls{smbo}-\gls{ei} is required
to escape local optima.
The proposed case study confirmed that the estimated exploration between
\gls{smbo}-\gls{pm} and  \gls{smbo}-\gls{ei}, as measured in distance to the nearest neighbor, decreases with increasing dimensionality.

Finally, we would like to end this paper with suggestions for future research.
First and foremost, the experiments clearly show that the \gls{pm} is a surprisingly competitive infill criterion. 
Future work on new infill criteria should include \gls{pm} in their benchmarks (e.g., as a baseline).
Portfolio methods that cover multiple infill criteria would likely profit from considering \gls{pm} in their framework.

Future work should reassess the importance of explorative \gls{smbo} in practical applications.
In this context, the sub-optimal performance of \gls{smbo}-\gls{ei} for dimensionalities of five or more is troubling and needs to be further investigated.

A consideration that was not covered in this work, is the global model quality.
A globally accurate model is not required for an optimization task that only searches for a single optimum.
However, there may be additional requirements in practice.
For instance, the learned model may have to be used after the optimization run
to provide additional understanding of the real-world process to practitioners or operators.
A model that is trained with data generated by a purely exploitative search
might fall short of this requirement, despite its ability to find a good solution.

\bibliographystyle{ACM-Reference-Format}
\bibliography{rehb20a} 

\end{document}